\documentclass{article}

\usepackage{booktabs}       
\newcommand{\mr}[2]{\multirow{#1}{*}{#2}}
\newcommand{\mc}[3]{\multicolumn{#1}{#2}{#3}}
\usepackage{amsfonts}       
\usepackage{nicefrac}       
\usepackage{microtype}      
\usepackage{array}
\newcolumntype{P}[1]{>{\centering\arraybackslash}p{#1}}
\usepackage{url}            

\usepackage{multicol}
\usepackage{amsmath,amsthm,amssymb,bm} 
\usepackage{xspace}
\usepackage{enumerate}
\usepackage{enumitem}
\usepackage{stmaryrd}
\usepackage{caption}
\usepackage{subcaption}
\usepackage{multirow}
\usepackage{times}
\usepackage{graphicx} 
\usepackage{sidecap}
\usepackage{wrapfig}
\usepackage[sort,numbers,square,comma]{natbib}
\usepackage{colortbl}
\usepackage{arydshln}
\usepackage{tikz}
\usepackage{listings}
\usepackage{xcolor}
\usepackage{makecell}
\usepackage{booktabs}
\usepackage{titletoc}
\usepackage{hyperref}
\usepackage{bbding}

\usepackage{hyperref}

\newcount\COMMENTs  
\usepackage{color}
\definecolor{darkgreen}{rgb}{0,0.5,0}
\definecolor{purple}{rgb}{1,0,1}
\newcommand{\comm}[2]{\ifnum\COMMENTs=1\textcolor{#1}{#2}\fi}

\newcommand{\ang}{\mbox{\normalfont\AA}}

\newcommand{\ocd}{OC20}
\newcommand{\dataset}{OC20}
\newcommand{\model}{ForceNet}

\newcommand{\ie}{\textit{i.e.}}
\newcommand{\eg}{\textit{e.g.}}

\DeclareSymbolFont{extraup}{U}{zavm}{m}{n}
\DeclareMathSymbol{\varheart}{\mathalpha}{extraup}{86}

\usepackage{amsmath,amsfonts,bm}

\def\eqref#1{equation~\ref{#1}}

\def\1{\bm{1}}

\DeclareMathAlphabet{\mathsfit}{\encodingdefault}{\sfdefault}{m}{sl}
\SetMathAlphabet{\mathsfit}{bold}{\encodingdefault}{\sfdefault}{bx}{n}

\newcommand{\R}{\mathbb{R}}

\usepackage[accepted]{icml2021}

\icmltitlerunning{\model{}: A Graph Neural Network for Large-Scale Quantum Calculations}

\begin{document}

\twocolumn[
\icmltitle{\model{}: A Graph Neural Network for Large-Scale Quantum Calculations}



\icmlsetsymbol{equal}{*}

\begin{icmlauthorlist}
\icmlauthor{Weihua Hu}{st}
\icmlauthor{Muhammed Shuaibi}{cmu}
\icmlauthor{Abhishek Das}{fb}
\icmlauthor{Siddharth Goyal}{fb}
\icmlauthor{Anuroop Sriram}{fb}
\icmlauthor{Jure Leskovec}{st}
\icmlauthor{Devi Parikh}{fb,ga}
\icmlauthor{C. Lawrence Zitnick}{fb}
\end{icmlauthorlist}

\icmlaffiliation{st}{Department of Computer Science, Stanford University}
\icmlaffiliation{cmu}{Department of Chemical Engineering, Carnegie Mellon University}
\icmlaffiliation{fb}{Facebook AI Research}
\icmlaffiliation{ga}{School of Interactive Computing, Georgia Institute of Technology}

\icmlcorrespondingauthor{Weihua Hu}{weihuahu@stanford.edu}
\icmlcorrespondingauthor{C. Lawrence Zitnick}{zitnick@fb.com}

\icmlkeywords{Graph Neural Networks}

\vskip 0.3in
]




\printAffiliationsAndNotice{} 

\begin{abstract}
With massive amounts of atomic simulation data available, there is a huge opportunity to develop fast and accurate machine learning models to approximate expensive physics-based calculations. The key quantity to estimate is atomic forces, where the state-of-the-art Graph Neural Networks (GNNs) explicitly enforce basic physical constraints such as rotation-covariance. However, to strictly satisfy the physical constraints, existing models have to make tradeoffs between computational efficiency and model expressiveness. 
Here we explore an alternative approach. By not imposing explicit physical constraints, we can flexibly design expressive models while maintaining their computational efficiency. 
Physical constraints are implicitly imposed by training the models using physics-based data augmentation.
To evaluate the approach, we carefully design a scalable and expressive GNN model, \model{}, and apply it to OC20~\citep{OC20}, an unprecedentedly-large dataset of quantum physics calculations. Our proposed \model{} is able to predict atomic forces more accurately than state-of-the-art physics-based GNNs while being faster both in training and inference. 
Overall, our promising and counter-intuitive results open up an exciting avenue for future research.
\end{abstract}

\section{Introduction}
\label{sec:intro}
Recently, massive physics-based data has been generated by ever-increasing scientific compute~\citep{OC20,nakata2019pubchemqc}. 
This provides a huge opportunity for Machine Learning (ML) approaches to efficiently and accurately model complex physical systems~\citep{sanchez2020learning,bapst2020unveiling,kipf2018neural,klicpera_dimenetpp_2020,battaglia2016interaction,gilmer2017neural,schutt2017schnet,klicpera2020directional}. 
An accurate ML model trained on large data can be used to perform inference orders-of-magnitude faster than the original 
physics-based calculations. 

Of particular practical interest is approximating atomic forces of quantum mechanical systems. This is because the underlying quantum calculations are expensive (several hours per system)~\citep{parr1980density}, and the resulting atomic forces can be used for diverse chemistry applications, such as structure relaxations, molecular dynamics, structural analyses, as well as transition state calculations.~\cite{behler2016perspective, del2019local, frederiksen2007inelastic, henkelman2000improved, henkelman2000climbing}

\begin{figure*}
\centering
\includegraphics[width=0.95\linewidth]{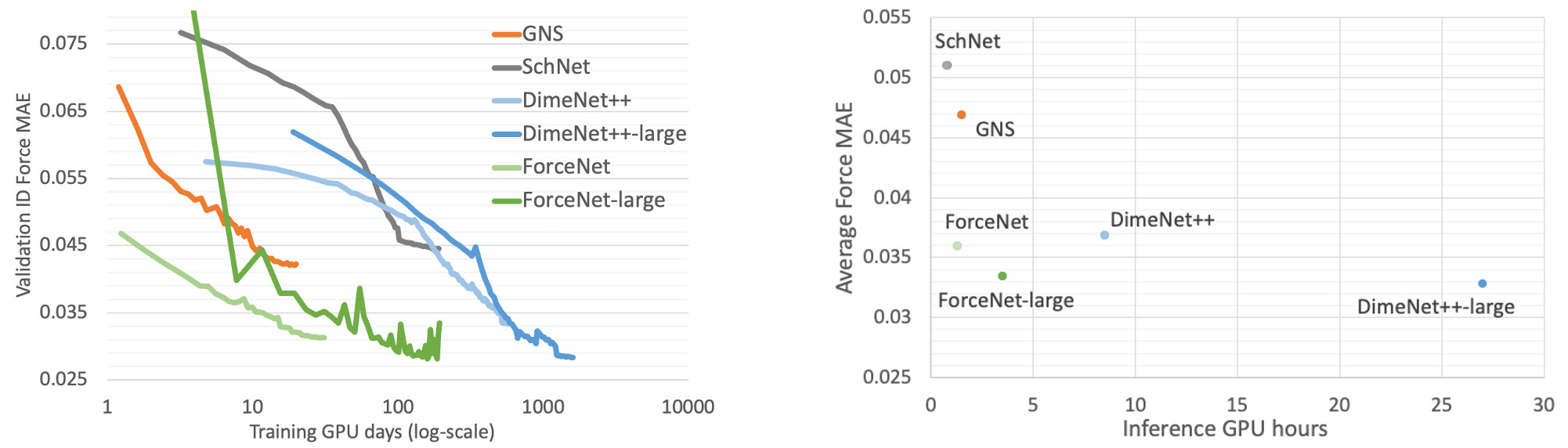}
\vspace{-0.4cm}
\caption{Comparison of S2F (atomic force prediction) performance across different models, while taking computational efficiency into account. \textbf{(Left)}: Comparison of validation learning curves, where $x$-axis is training GPU days in the \emph{log-scale} (lower left is better). \textbf{(Right)}: Comparison of validation performance and inference time in GPU hours, measured over the in-distribution validation set (lower left is better). GPUs with the same specs are used for fair comparison (details in Section~\ref{subsec:model_config}).}
\label{fig:main_comp_efficiency}
\vspace{-0.3cm}
\end{figure*}

The state-of-the-art approach to predicting atomic forces is physics-based message-passing Graph Neural Networks (GNNs)~\citep{gilmer2017neural}, with the representative models being SchNet~\citep{schutt2017schnet} and DimeNet~\citep{klicpera_dimenetpp_2020,klicpera2020directional}. These GNNs first predict the energy of the entire system in a rotation-invariant manner, and then predict the per-atom forces by taking the derivative of the energy with respect to the atomic positions. 
By the architecture's design, these GNNs produce forces that obey the basic physical rules of rotation-covariance and energy-conservation.

However, designing effective GNNs, while satisfying these physical rules is highly non-trivial. For instance, SchNet~\citep{schutt2017quantum} is computationally efficient, but the model only uses atomic distances in its message passing in order to ensure rotation-invariance of its energy prediction. Consequently, SchNet fails to capture the 3D structure explicitly, resulting in sub-optimal generalization performance.
The recent DimeNet and DimeNet++~\citep{klicpera2020directional,klicpera_dimenetpp_2020} additionally capture bond angle information in its message passing, but this comes with the cost of expensive message computations involving atom triplets to ensure the rotation-invariance. As a result, DimeNet necessitates tremendous compute to scale to a massive dataset (Figure~\ref{fig:main_comp_efficiency} (left))---1600 GPU days to train DimeNet++-large.
Moreover, even DimeNet is unable to model an important physical feature of torsion angles~\citep{leach2001molecular}, failing to capture the full 3D information in its message passing.

Here we explore an alternative approach, building on the recent framework of Graph Network-based Simulators (GNS)~\citep{sanchez2020learning,bapst2020unveiling}. 
Specifically, by not explicitly imposing physical constraints in the model architecture, we can flexibly design expressive GNN models and use the full 3D atomic positions in a scalable manner.
In exchange, the predicted forces are translation-invariant but no longer rotation-covariant. 
As we demonstrate empirically, this issue can be alleviated by training models on a massive dataset with rotation data augmentation. In other words, we impose physical constraints to the model \emph{implicitly through physics-based data} rather than explicitly through architectural constraints. Our model also does not explicitly enforce energy conservation. However, this removes the memory-intensive calculations in physics-based GNNs, \ie, compute forces through energy gradients that require second-order derivatives to optimize.

To realize our approach, we carefully design a GNN model that accurately captures 3D atomic structure in a scalable and flexible manner.
The resulting model is \model{} that uses an expressive message passing architecture with carefully-chosen basis and non-linear activation functions. 

\begin{figure}
	\begin{center}
		\includegraphics[width=0.88\linewidth]{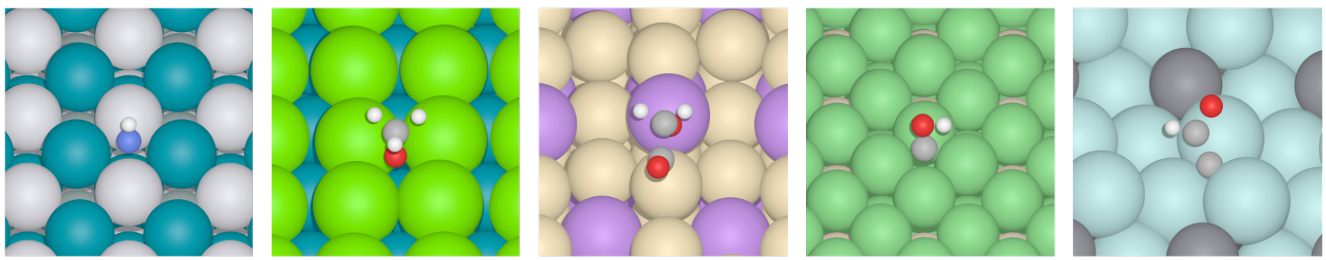}
	\end{center}
	\vspace{-0.2cm}
	\caption{Illustration of sampled systems from the OC20 dataset~\citep{OC20}. Each system consists of adsorbate (the small molecule on the surface) and catalysis (the large grid-like molecule sitting below the adsorbate), and is repeated in the direction of the horizontal axes infinitely. Our \model{} aims to efficiently predict per-atom forces.}
	\label{fig:catalysis}
	\vspace{-0.5cm}
\end{figure}

We evaluate \model{} on OC20~\cite{OC20}, a recently-introduced large-scale dataset of quantum physics calculations with 200+ million large atomic structures (20--200 atoms) useful for discovering new catalysts for energy applications~\cite{Seheaad4998, Jouny2018,zitnick2020introduction} (Figure~\ref{fig:catalysis}).
The dataset was constructed with an unprecedented 70 million CPU hours of compute performing Density Functional Theory (DFT)-based quantum calculations~\citep{parr1980density}---more than 20 times the compute as compared to the conventional quantum physics datasets of QM9~\citep{ramakrishnan2014quantum} and Alchemy~\citep{chen2019alchemy}, making it ideal for scalable deep learning approaches. 
Moreover, unlike many existing quantum physics datasets, OC20 provides non-equilibrium structures of molecules, \ie, 3D structure with non-zero atomic forces, making it a good testbed of our model.

Even without any explicit physical constraints, \model{} is able to achieve higher accuracy than physics-based GNNs when trained with comparable computational resources (Figure~\ref{fig:main_comp_efficiency} (left)). Moreover, \model{} (resp.~\model{}-large) achieves prediction errors that are comparable to the state-of-the-art DimeNet++~(resp.~DimeNet++-large) with 6 (resp.~8) times less inference time (Figure~\ref{fig:main_comp_efficiency} (right)).
Finally, compared to DimeNet++, \model{}-large achieves more accurate force prediction, while being faster both in training and inference (Figure~\ref{fig:main_comp_efficiency} (left) and (right)).

To understand the \model{}'s design choices, we perform extensive ablation studies on each architectural component of \model{}. We find that the expressive edge-level computation for accurately modeling 3D atomic interactions contributes most to the model performance. Overall, we demonstrate that even without explicit physical constraints, a scalable expressive GNN provides promising performance in modeling complex physical systems, opening up an exciting avenue for future research.

\section{Related Work}
\label{sec:related}

ML for approximating quantum physics calculations has been extensively studied in the literature~\citep{christensen2020role, chmiela2017machine, chmiela2018towards, khorshidi2016amp,schutt2017schnet,ramakrishnan2014quantum,klicpera_dimenetpp_2020,klicpera2020directional,gilmer2017neural,chen2019alchemy}. These studies have been either on relatively small-scale datasets (in the order of 100K structures), small molecule size (10--30 atoms), or equilibrium structures (3D structures with all-zero atomic forces). In contrast, our focus is on a large-scale dataset (in the order of 100M structures), larger molecule size (20--200 atoms), and non-equilibrium structures (3D structures with non-zero atomic forces), making the model scalability and expressiveness especially important.

Message-passing GNNs~\citep{gilmer2017neural} have been particularly effective in modeling quantum physical systems.
Below, we review GNNs and their two major approaches to modeling atomic forces.

\vspace{-0.2cm}
\paragraph{Message-passing GNNs.}
\model{} is based on message passing GNNs that iteratively update node embeddings based on messages passed from neighboring nodes. In its most general form, the message function depends on the two node embeddings as well as edge features. Many GNN variants fall under this framework~\citep{kipf2016semi,velivckovic2017graph,xu2018how,hamilton2017inductive}. The GNN-FiLM~\citep{brockschmidt2019gnn} uses an embedding of the target node to modulate the message from the source nodes, which is closely related to our expressive message passing architecture. However, most existing message-passing GNNs, including GNN-FiLM, are designed for homogeneous graphs without edge features. Consequently, edge features, which are central to our problem, are often incorporated in an adhoc manner and can be ineffective at approximating complex atomic interactions.

\vspace{-0.2cm}
\paragraph{Force-centric Models.} 
\model{} builds on the force-centric GNS framework~\citep{sanchez2020learning,bapst2020unveiling,park2020accurate}. Here a model's primary output is per-atom forces (thus, force-centric). 
The GNS framework follows three steps to predict forces: (1) A graph is constructed from 3D points, (2) an encoder GNN is applied to the graph to obtain node embeddings, and (3) a decoder is applied to the node embeddings to predict the per-node forces.
The GNS framework has been applied to relatively simple physical systems such as fluids, rigid body, and glassy systems, where ground-truth calculations are already cheap and can be performed on-the-fly during training.
Compared to these domains, using ML to approximate expensive quantum physics calculations is more practically impactful and challenging. Whether GNS is effective in the practically-relevant applications remains largely open.
As we demonstrate empirically, the off-the-shelf GNN model used in GNS fails to accurately predict the quantum mechanical forces, necessitating more careful design of model architectures. 

\vspace{-0.2cm}
\paragraph{Energy-centric Models.} 
The majority of GNN models developed for quantum physics calculations fall under the energy-centric simulation framework, in which a model's primary output is the energy of the entire atomic system (hence, energy-centric).
Rotationally-invariant GNNs are used to predict the energy. 
Atomic forces are then predicted implicitly through negative gradients of energy with respect to the atomic positions, which can be directly regressed to the ground-truth forces using the second-order derivatives.
The architecture guarantees that the force-field obeys the basic physical rules of rotation-covariance and energy-conservation.  
 
Many advanced GNN architectures have been proposed under the energy-centric framework, such as SchNet~\citep{schutt2017schnet}, DimeNet~\citep{klicpera2020directional}, and its recent improvement, DimeNet++~\citep{klicpera_dimenetpp_2020}. 
As we discuss in the introduction, these models are either computationally expensive (DimeNet involves message passing over triplets of atoms) or unable to explicitly capture angular information among a set of atoms (Schnet's message passing only depends on atomic distances).
Although DimeNet captures angular information, it is still restricted to bond angles, and torsion angles are not captured explicitly~\citep{leach2001molecular}. Capturing the full angular information in a rotationally-invariant manner would require even more computation, \eg, message passing over atom quadruplets.

Our scalable expressive message passing architecture is built from SchNet's \emph{continuous filter convolution} architecture, where we make an important extension to resolve a number of critical issues when adopting it to the force-centric GNS framework (see Section \ref{subsec:condfilter} for details).

\section{\model{}}
\label{sec:model}
Here we introduce \model{} by describing its model architecture as well as the effective data augmentation strategy to encourage rotation covariance of \model{}'s predictions. 
The input to \model{} is an atomic structure, \ie, a set of atoms and their 3D spatial positions (Figure~\ref{fig:catalysis}). The output is a 3D vector for each node, representing the predicted $(x, y, z)$ atomic force.

\subsection{Model architecture}

\model{} represents atoms as nodes in a GNN and the atomic interactions as edges. The node input features specify the atom's atomic number and other properties (9-dimensional vector adapted from~\citet{xie2018crystal}). Edges in the GNN are constructed from a radius graph of neighboring atoms~\citep{schutt2017schnet,sanchez2020learning}. Let $\mathcal{N}_t(c)$ denote a set of neighboring atoms that are within the cutoff-distance $c$ away from the target atom $t$. On average an atom has 35 neighbors. A directed edge from source atom $s$ to target atom $t$ is drawn for $s \in \mathcal{N}_t(c)$. Let $\bm{d}_{st} \in \R^3$ be their relative displacement, \ie, a vector pointing from atom $s$ to atom $t$. 

\model{} follows the encoder-decoder architecture of the GNS framework~\citep{sanchez2020learning,battaglia2016interaction,kipf2018neural}. The encoder uses scalable iterative message passing to compute node embeddings $\bm{h}_t$ that capture the $3$D structure surrounding each atom, and the decoder uses an Multi-Layer Perceptron (MLP) to directly predict per-atom forces from these embeddings.
The encoder updates $\bm{h}_t$ as:
\begin{align}
\label{eq:forcenetmessage}
\bm{h}_t^{(k+1)} = \bm{F}_n\left(\bm{m}_t + \sum_{s\in\mathcal{N}_t}\bm{m}_{st}\right) + \bm{h}_t^{(k)},
\end{align}
where the messages $\bm{m}_{st}$ and $\bm{m}_t$ are summed and passed through the function $\bm{F}_n: \R^D \to \R^D$ that is a 1-hidden-layer MLP with batch normalization~\citep{ioffe2015batch}. The dimensionality of the node and hidden layer features is $D$. Equation (\ref{eq:forcenetmessage}) follows standard GNN embedding update formulations \citep{gilmer2017neural} with the addition of a residual connection, $\bm{h}_t^{(k)}$ ~\citep{he2016deep}. We define the pairwise messages $\bm{m}_{st}$ and self message $\bm{m}_t$ in Section~\ref{subsec:condfilter}. 

The decoder is computed using the last layer $K$'s node embeddings $\bm{h}_t^{(K)}$, $\bm{f}_t = \bm{F}_f(\bm{h}_t^{(K)})$ where $\bm{f}_t$ is the 3D force of atom $t$, and $\bm{F}_f$ is a 1-hidden-layer MLP with batch normalization.

The critical aspect of \model{} is its encoder and specifically the scalable message computation that effectively captures the non-linear and complex 3D atomic interactions to predict the atomic forces. In the following, we present three key architectural components in our message computation.

\begin{figure}
\centering
\includegraphics[width=0.95\linewidth]{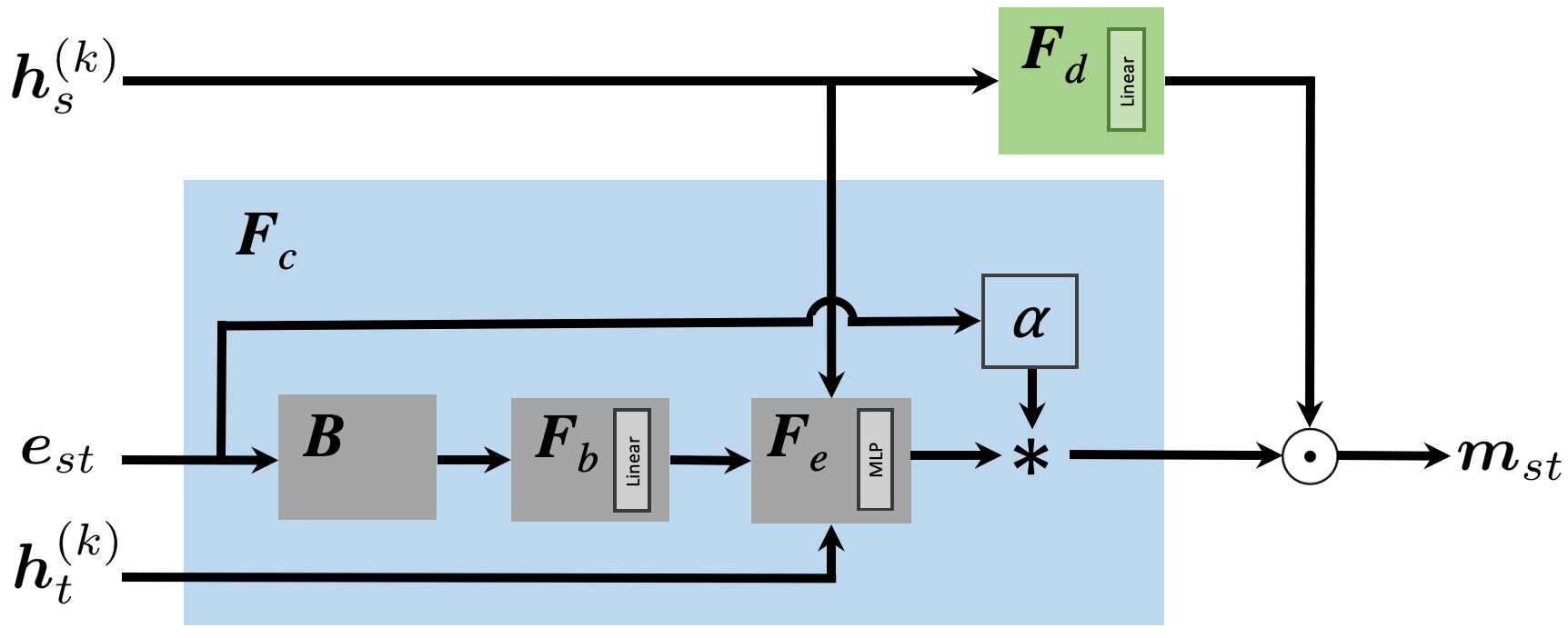}
\caption{Model diagram for messages $\bm{m}_{st}$ (from atom $s$ to atom $t$) used by \model{} in Eqns. (\ref{eq:model_message}) and (\ref{eq:conditional_filter}). The key components are (a) the expressive conditional filter $\bm{F}_c$ that is dependent on full edge feature $\bm{e}_{st}$ (complete 3D relative placement information) as well as source and target node embeddings, $\bm{h}^{(k)}_s$ and $\bm{h}^{(k)}_t$, (b) the basis function $\bm{B}$ over the edge feature that helps the network to accurately capture atomic interactions, and (c) the smooth curved non-linearity of the Swish activation.}
\label{fig:model}
\end{figure}

\subsubsection{Conditional Filter Convolution}
\label{subsec:condfilter}
We first present conditional filter convolution, a simple yet effective extension of SchNet's scalable continuous filter convolution~\citep{schutt2017schnet}. 
The original continuous filter convolution uses the distance information between neighboring atoms to compute the filter, which is then applied to the embeddings of source atoms. 
However, this approach has a series of limitations, especially when transitioning from an energy-centric to a force-centric model. 

First, to ensure rotation-invariant energy prediction, SchNet's continuous filter only uses the atom distance as the input edge feature. Hence, the angular information is lost in the message passing. More crucially, the resulting node embeddings are rotation-invariant, but forces need to rotate together with a molecular system. Furthermore, the filter does not depend on the source and target atoms. However, changes in their atom types can result in significant differences in forces between the atoms even if they are placed at a similar distance, because of their varying electronic properties~\cite{gibbs1998power}. Hence, the filter may not be expressive enough to model complex non-linear atomic interactions.

We resolve these issues by using our $E$-dimensional edge feature $\bm{e}_{st}$ (described below) that encodes rotation-covariant directional information, and conditioning on both the source $\bm{h}_s^{(k)}$ and target $\bm{h}_t^{(k)}$ node information:
\begin{align}
\label{eq:model_message}
\bm{m}_{st} = \bm{F}_c(\bm{h}_s^{(k)}, \bm{e}_{st}, \bm{h}_t^{(k)}) \odot \bm{F}_d(\bm{h}_s^{(k)}),
\end{align}
where $\bm{F}_c: \R^D \times \R^E \times \R^D \to \R^D$ is the conditional filter, and $\bm{F}_d: \R^D \to \R^D$ is a learnable linear function that transforms the source node embeddings before the filter is applied. The edge feature is defined as $\bm{e}_{st} \equiv {\rm Concat} (\bm{n}_{st}, \bm{p}_{st}/c) \in \R^7$, where $\bm{n}_{st} \equiv \bm{d}_{st} / \|\bm{d}_{st}\|$ is a normalized directional vector and $\bm{p}_{st} \in \R^{4}$ is a list of four atomic distances $\|\bm{d}_{st}\|, \|\bm{d}_{st}\| - a_s, \|\bm{d}_{st}\| - a_t, \|\bm{d}_{st}\| - a_s - a_t$ that take into account the atomic radii~\citep{slater1964atomic} $a_s$ and $a_t$ of atoms $s$ and $t$, respectively.

The conditional filter $\bm{F}_c$ combines the raw edge features $\bm{e}_{st}$ with the node embeddings to encode the interactions between atoms $s$ and $t$ (Figure \ref{fig:model}) and is defined as :
\begin{align}
\label{eq:conditional_filter}
    &\bm{F}_c(\bm{h}_s^{(k)}, \bm{e}_{st}, \bm{h}_t^{(k)}) = \nonumber \\
    &\qquad \alpha(\|\bm{d}_{st}\|) \cdot \bm{F}_{e}\left(\bm{h}_s^{(k)}, \bm{F}_b(\bm{B}\left(\bm{e}_{st})\right), \bm{h}_t^{(k)} \right) ,
\end{align}
where $\alpha(x)=\cos(\pi x / 2c)$ is a scalar that decays to zero as $\|\bm{d}_{st}\|$ approaches the distance cutoff $c$. $\bm{F}_{e}$ is a 2-hidden-layer MLP with the hidden size of $D$ and the three input vectors are concatenated as input. $\bm{B}: \R^E \to \R^B$ is the basis function we discuss in the next section, and $\bm{F}_b: \R^B \to \R^D$ is a learnable linear function that maps the $B$ dimensional vector to a $D$ dimensional vector for input to $\bm{F}_{e}$. The parameters for $\bm{F}_b$ are shared across layers, while no other parameters are shared across layers. Finally, the self message $\bm{m}_t$ is defined by applying an element-wise product between learnable filter $\bm{v} \in \R^D$ and $\bm{F}_d$, \ie, $\bm{m}_t = \bm{v} \odot \bm{F}_d(\bm{h}_t^{(k)})$. 

{\bf Comparisons with the existing GNS model.} Notice that the filter $\bm{F}_c$ itself already contains the source node information $\bm{h}_s^{(k)}$, and may be directly aggregated, as done in the off-the-shelf GNS models~\citep{sanchez2020learning,bapst2020unveiling}.
Nonetheless, we empirically find that explicitly applying $\bm{F}_c$ on $\bm{F}_d$ through the element-wise dot product in Equation~(\ref{eq:model_message}) significantly improves the performance.
There are also other subtle but important differences between \model{} and the exising GNS model, such as the use of basis functions (Section~\ref{subsec:basis}) and the choice of non-linear activations (Section~\ref{subsec:activation}).
We empirically show that these careful architecture design choices in \model{} contribute to the significant performance improvement over the existing GNS model.

{\bf Enforcing rotation-covariance.}
Note also that the use of $\bm{n}_{st}$ in $\bm{e}_{st}$ results in the model being not necessarily rotation invariant nor covariant.
In Section \ref{subsec:train}, we propose to encourage the physical constraint by training models with rotation data augmentation.
In Section \ref{subsec:rotcov}, we empirically demonstrate the effectiveness of this strategy in encouraging the rotation-covariance of \model{}'s predictions.

\subsubsection{Basis Functions}
\label{subsec:basis}

An important aspect of $\bm{F}_c$ is the choice of basis function $\bm{B}: \R^E \to \R^B$ that transforms the raw distance features $\bm{e}_{st}$ into ones that are more discriminative. Several choices of basis functions have been proposed, such as a Gaussian over 1D distances~\citep{schutt2017schnet} and a spherical Bessel function over the joint 2D space of the edge distance and angle~\citep{klicpera2020directional}. We extend these ideas to capture the full 3D positional differences between atoms, and systematically study the effectiveness of different basis functions in the context of a force-centric model.

Each of our basis functions $\bm{B}$ maps the raw edge features $\bm{e}_{st}$ presented in Section~\ref{subsec:condfilter} into a $B$ dimensional vector, where $B$ varies based on the basis function used. $B$ is typically much larger than $E=7$ to aid in capturing subtle differences in atom positions.


{\bf Identity:} $\bm{B}_{\rm id}(\bm{x}) = \bm{x}$. The baseline is to use the edge features $\bm{e}_{st}$ directly. 

{\bf Linear + Act:} $\bm{B}_{\rm linact}(\bm{x}) = {\rm g}(\bm{W}\bm{x} + \bm{b})$, where ${\rm g}(\cdot)$ is the non-linear activation function, and $\bm{W}$ and $\bm{b}$ are the learnable parameters. When followed by the linear layer $\bm{F}_b$, this is equivalent to applying an 1-hidden-layer MLP over the edge features $\bm{e}_{st}$.

{\bf Gaussian:} $\bm{B}_{\rm gauss}(\bm{x}) = [b_1,\ldots,b_J]$, where $b_j$ is the output of the $j$-th basis function $b_j(x) = {\rm exp}^{ \left( x - \mu_j \right)^2 / (2 \cdot \sigma^2)}$. The Gaussian means are evenly distributed on the interval between 0 and 1, \ie, $\mu_j = j / (J-1)$ and the standard deviation $\sigma = 1/(J-1)$. All values of $x$ are normalized to lie between 0 and 1. $\bm{B}_{\rm gauss}$ is applied to each dimension of $\bm{e}_{st}$, resulting in a $B = J \times E$ vector.

{\bf Sine:} $\bm{B}_{\rm sin}(\bm{x}) = [b_1,\ldots,b_J]$, where $b_j$ is the output of the $j$-th basis function $b_j(x) = \sin(1.1^j x)$. The design is based on function approximation using the Fourier series. In our experiments, we find that using only the sinusoidal component of the Fourier series is sufficient. $\bm{B}_{\rm sin}$ is applied to each dimension of $\bm{e}_{st}$, resulting in an $B = J \times E$ vector. 

{\bf Spherical harmonics:} $\bm{B}_{\rm sph}(\bm{e}_{st}) = \bm{Y}_L(\theta, \phi) \bm{R}(\bm{p}_{st})^{\top}$ where $\bm{Y}_L$ is the list of Laplace's spherical harmonics~\citep{macrobert1947spherical} used to encode the angular information and $\bm{R}(\bm{p}_{st})$ encodes the distance. We use spherical harmonic functions up to degree $L$, which gives us $L^2$ orthogonal basis in total. The angles $\theta$ and $\phi$ can be directly computed from $\bm{n}_{st}\in \R^3$ in $\bm{e}_{st}$. $\bm{R}$ uses a linear combination of the above sine basis functions computed from $\bm{p}_{st} \in \R^4$ in $\bm{e}_{st}$ (thus, $4J$ basis functions in total) to encode distance information. Specifically, $\bm{R}(\bm{p}_{st}) = \bm{W}_{\rm rad} \bm{B}_{\rm sin}(\bm{p}_{st}) + \bm{b}_{\rm rad} \in \R^S$, where $\bm{W}_{\rm rad}$ and $\bm{b}_{\rm rad}$ are learnable parameters. $\bm{B}_{\rm sph}$ is flattened into a vector before being passed into $\bm{F}_{b}$. The dimensionality of $\bm{B}_{\rm sph}$ is $B=SL^2$, where we set $S$ to be the dimensionality of $\bm{p}_{st}$. 

\subsubsection{Expressive Non-Linearity in MLPs}
\label{subsec:activation}
Our final key component is simple but crucial: the choice of non-linear activation function plays a central role in modeling complex non-linearities of atomic intereactions. 
The ReLU activation~\citep{glorot2011deep} is widely used in many deep learning models, including the existing GNS model~\citep{sanchez2020learning,bapst2020unveiling}.
However, ReLU may not be ideal in modeling atomic forces, since it results in outputs being modeled as piece-wise linear hyper-planes with sharp boundaries. Ideally, we desire a smooth and expressive non-linear activation function.

We explored a wide array of choices for activation functions, such as Tanh, Leaky-ReLU, SoftPlus~\citep{dugas2001incorporating}, Shifted SoftPlus~\citep{schutt2017schnet}, and Swish~\citep{ramachandran2017swish}. In our experiments, we find Swish, \ie, ${\rm act}(x) = x\cdot {\rm sigmoid}(x)$, to perform particularly well. As illustrated in Figure~\ref{fig:basis}, Swish provides a smoother output landscape and has non-zero activation for negative inputs. As we demonstrate in Section \ref{subsubsec:basis} and Figure~\ref{fig:basis}, the replacement of ReLU with Swish consistently and significantly improves the predictive accuracy while maintaining scalability across all choices of basis functions. 

\vspace{-0.1cm}
\subsection{Rotation Data Augmentation}
\vspace{-0.1cm}
\label{subsec:train}
We apply random rotation data augmentation to encourage the rotation-covariance of the model's predictions. 
Specifically, we randomly rotate the entire system and per-atom forces by the same degree, and let \model{} predict the rotated forces based on the rotated system.

Rotation augmentation is particularly effective when the physical systems of interest have a canonical axis. Such systems are prevalent in many real applications. For instance, most of the material-type molecular systems have the canonical axis that is vertical to the material surface (Figure~\ref{fig:catalysis}). 
At the macroscopic level, most of the physical systems have the natural canonical axis pointing towards the direction of gravity of the earth. 
Given such a canonical axis, rotation augmentation only needs to be applied along a single axis, making learning more data efficient.

In this work, we apply \model{} to atomic structures that do have a canonical vertical axis perpendicular to the material surface. We explicitly make use of this property and only apply rotation augmentation along this vertical axis. Empirically, \model{} trained with large data and our data-efficient rotation augmentation strategy learns to closely approximate rotation-covariance, as we demonstrate in Section~\ref{subsec:rotcov}.


\section{Experiments}
\label{sec:experiments}
In this section, we evaluate \model{}'s performance in predicting atomic forces. We do so by applying the model to OC20~\citep{OC20}, a massive dataset on quantum physics calculations on non-equilibrium atomic structures relevant to catalysis discovery.

Throughout this section, we normalize for computational time when comparing models' predictive performance, \ie, we compare models with similar computational cost in training and inference. This is crucial because simply using more computational resources to train larger models is shown to lead better results in OC20 tasks~\citep{OC20}. However, training time of most existing models is already more than 100 GPU days\footnote{Defined as the number of GPUs times the number of days the GPUs are used.} and even goes up to 1600 GPU days, making it harder to further scale up without improving the models' computational efficiency. Moreover, fast model inference is crucial for the application of catalyst material discovery, where an ML model needs to make predictions over an enormous number of potential candidates~\citep{zitnick2020introduction}.

\subsection{Task Descriptions and Evaluation Metrics}
OC20 dataset~\citep{OC20} contains 200M+ non-equilibrium 3D atomic structures from 1M+ atomic relaxation trajectories. Each structure is associated with the per-structure energy and per-atom forces. Figure~\ref{fig:catalysis} shows an illustration of 3D structures.

OC20 provides a variety of prediction tasks relevant to catalyst discovery for renewable energy applications. Our main focus is on the atomic force prediction task, called S2F (\textbf{S}tructure to \textbf{F}orces).
Following the baseline setting in S2F~\citep{OC20}, we train our models on the 130M training structures that are on relaxation simulation trajectories.
In this work, we focus on the S2F task. In Appendix~\ref{app:IS2RS}, we also provide preliminary results on a simulation task by directly applying our S2F models.


We evaluate models on four validation datasets that test different levels of model generalization: In Domain (ID), Out of Domain Adsorbate (OOD Adsorbate), OOD Catalyst, and OOD Both (both the adsorbate and catalyst's material are not seen in training). Each split contains 1M examples. 

Following~\citet{OC20}, the Mean Absolute Error (MAE) of forces on free atoms is evaluated for each validation set. Here the free atoms represent atoms that are close to the material surface and are free to move during atomic relaxation simulation (Figure~\ref{fig:slab} in Appendix~\ref{app:dataset}). We use the ``average force MAE'' to represent the MAE of forces averaged over the four validation sets.

\subsection{Model Settings and Computational Time}
\label{subsec:model_config}
Below, we describe hyper-parameter settings of \model{} and baseline models, along with the computational time (in GPU days) to train these models. Table~\ref{tab:comp} shows the summary of training time, inference time, and sizes of different models. 
All the training is run under Tesla V100 Volta. The inference time is measured on the validation ID set under GeForce RTX 2080, where the largest possible batch size is used for each model.

\begin{table*}[t]
    \centering
        \caption{Comparison of \model{} to existing GNN models.
    We mark as bold the best performance and close ones, \ie, within 0.0005 MAE, which according to our preliminary experiments, is a good threshold to meaningfully distinguish model performance. Training time is in GPU days, and inference time is in GPU hours. 
    Median represents the trivial baseline of always predicting the median training force across all the validation atoms. 
    }
    \label{tab:comp}
    \renewcommand{\arraystretch}{1.0}
    \setlength{\tabcolsep}{5pt}
    \resizebox{0.97\linewidth}{!}{
    \begin{tabular}{lrrrrrcccccc}
      \toprule
         \mr{2}{\textbf{Model}} & \textbf{Hidden} & \textbf{\#Msg} & \mr{2}{\textbf{\#Params}} & \textbf{Train}  & \textbf{Inference} & \mc{5}{c}{\textbf{Validation Force MAE (eV/$\ang$)}} \\
        & \textbf{dim} & \textbf{layers} & & \textbf{time} & \textbf{time} & ID & OOD Ads. & OOD Cat. & OOD Both & {\bf Average} \\
      \midrule
        Median & -- & -- & -- & & &  0.0810 & 0.0799 & 0.0799 & 0.0943 & 0.0838\\
      \midrule
        GNS  & 768 & 5 & 12.5M & 20d & 1.5h & 0.0421  & 0.0466 & 0.0430 & 0.0559 & 0.0469 \\
        SchNet  &  1024 & 5 & 9.1M & 194d & 0.8h &  0.0443 & 0.0514 & 0.0465 & 0.0618 & 0.0510 \\
        DimeNet++ & 192 & 3 & 1.8M &  587d & 8.5h & 0.0332 & 0.0366 & 0.0344 & 0.0436 & 0.0369 \\
        DimeNet++-large  & 512 & 3 & 10.7M & 1600d & 27.0h  & \textbf{0.0281} & \textbf{0.0318} & \textbf{0.0315} & \textbf{0.0396} & \textbf{0.0328} \\
        \midrule
        \bf \model{} & 512 & 5 & 11.3M & 31d & 1.3h & 0.0313 & 0.0355 & 0.0334 & 0.0439 & 0.0360 \\
        {\bf \model{}-large} & 768 & 7 & 34.8M & 194d & 3.5h &  \textbf{0.0281} &  \textbf{0.0320} & 0.0327 & 0.0412 & 0.0335 \\
      \bottomrule
    \end{tabular}
    }
    \vspace{-0.35cm}
\end{table*}

\begin{figure*}
\centering
\begin{minipage}{0.26\textwidth}
\centering
    \captionof{table}{Ablations on the architecture of \model{}.}%
   \label{tab:ablation_message}
    \resizebox{\linewidth}{!}{
    \begin{tabular}{lc}
      \toprule
         \mr{2}{\textbf{Ablation}} &  \textbf{Average Force} \\
          & \textbf{MAE (eV/$\ang$)} \\
      \midrule
         {\bf \model{}}  & {\bf 0.0360}  \\
         (1) Only-dist & 0.0699  \\
         (2) No-atomic-radii & 0.0368  \\
         (3) No-node-emb & 0.0410  \\
         (4) Only-${\bm F}_c$ & 0.0378  \\
         (5) Edge-linear-BN & 0.0427  \\
         (6) Node-linear-BN & {\bf 0.0364} \\
         (7) No-${\bm m}_t$ & {\bf 0.0364} \\
      \bottomrule
    \end{tabular}}
 \end{minipage}
 \hspace{0.5cm}
 \begin{minipage}{0.37\textwidth}
 \includegraphics[width=0.95\linewidth]{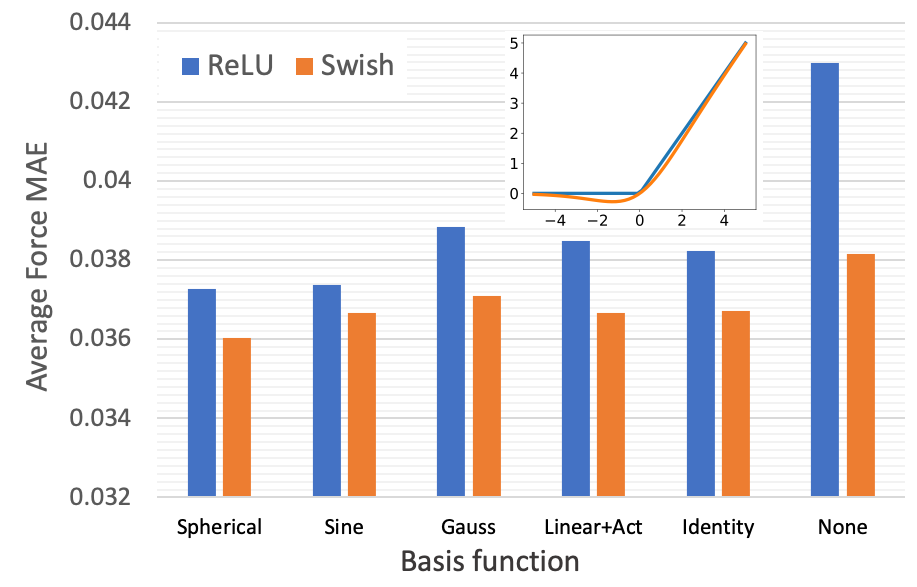}
 \vspace{-0.5cm}
\caption{Ablations on basis and activation functions in \model{}.}
\label{fig:basis}
\end{minipage}
\hspace{0.5cm}
\begin{minipage}{0.29\textwidth}
    \captionof{table}{Ablation of model scaling in terms of : (a) hidden dimensionality, (b) number of message passing layers, and (c) training batch size. Training time is roughly proportional to (b) and (c) and quadratic in (a).}
    \label{tab:scale}
    \resizebox{\linewidth}{!}{
    \begin{tabular}{rrrcccccc}
      \toprule
        \textbf{Hidden} & \textbf{\#Msg} & \textbf{Batch}  &  \textbf{Average Force} \\
        \textbf{dim}  & \textbf{layers} & \textbf{size} & {\bf MAE (eV/$\ang$)}  \\
      \midrule
         512 & 5 & 256 & 0.0360 \\
        768  & 5 & 256 & 0.0352 \\
        512 & 7  & 256 & 0.0355 \\
        768 & 7  & 256 & 0.0352 \\
        {\bf 768} &\textbf{7} & \textbf{512} & {\bf 0.0345}\\
      \bottomrule
    \end{tabular}
    }
\end{minipage}
\vspace{-0.3cm}
\end{figure*}

\paragraph{\model{}.}
We use the spherical function and Swish activation as the default basis and activation functions, since this combination consistently provides the best results (Figure \ref{fig:basis}).
The default model size has 5-layers of message passing and 512-dimensional hidden channels, and the training batch size is set to 256.
We also consider a larger variant, \model{}-large, that uses 7-layer message passing, 768-dimensional hidden channels, and the batch size of 512.
Training \model{} and \model{}-large takes 31 and 194 GPU days, respectively. 

During training, we apply the data-efficient rotation augmentation strategy presented in Section~\ref{subsec:train}.
We also find it useful to train on both free and fixed atoms, even though the evaluation is only on the free atoms.
Specifically, we give a small relative weight of 0.05 to the loss of fixed atoms during training, which is ablated in Table~\ref{tab:train_strategy} of Appendix \ref{app:ablation}.
Further implementation details and hyper-parameters are provided in Appendix \ref{app:hyperparam}. 

\paragraph{Baseline models.}
We compare \model{} against the following three strong baseline GNN models.
\vspace{-0.2cm}
\begin{itemize}[leftmargin=3mm]
    \item \textbf{SchNet}~\citep{schutt2017schnet} is a energy-centric GNN that uses scalable atom-pair-based message passing; hence, a relatively large model size (5-layer message passing with 1024-dimensional hidden channels) can be trained with 194 GPU hours, which is comparable to \model{}-large.
    \vspace{-0.1cm}
    \item \textbf{DimeNet++}~\citep{klicpera_dimenetpp_2020} is a recent improvement of DimeNet~\citep{klicpera2020directional} and is also an energy-centric GNN. It uses atom-triplet-based message passing to capture angular information, which makes it computationally expensive.
Even training DimeNet++ of a relatively-small model size (3-layer message passing with 192-dimensional hidden channels) requires 587 GPU days---18.9 and 3.0 times more expensive than \model{} and \model{}-large, respectively.
Training DimeNet++-large (3-layer message-passing with 512-dimensional hidden channels) takes a significant 1600 GPU days of compute, being 51.6 and 8.2 times more expensive than \model{} and \model{}-large, respectively.
    \vspace{-0.1cm}
    \item \textbf{GNS model}~\citep{sanchez2020learning} is a scalable force-centric model that directly predicts atomic forces.
We make its model size (in terms of the number of parameters) comparable to \model{}. The training takes 20 GPU days, which is $1.6\times$~faster than \model{}. However, as we will see, the performance of \model{} is better even if \model{}'s training is truncated at 20 GPU days.
\end{itemize}
\vspace{-0.2cm}
All the results of SchNet and DimeNet++ are directly adopted from the OC20 paper~\citep{OC20}. These energy-centric models are trained only on atomic forces, although in principle, they can be aso trained on per-system energy. \citet{OC20} report that training on forces and energy seperately achieves better performance on each task compared to joint training.
For the GNS model, we reproduce the original model architecture ourselves based on the feedback from the original author of GNS~\citep{sanchez2020learning}. Refer to Appendix \ref{app:gns} for implementation details. On the GNS model, we apply the same training strategies as \model{}.

\begin{table}[t]
    \centering
    \caption{Analysis of how training data and rotation augmentation affect the stability of \model{}'s prediction against rotation. 
    1000 validation ID structures are sampled and randomly rotated 100 times along the vertical axis. For each rotated structure, \model{} predicts per-atom forces that are then rotated back to compare with the originally-predicted forces. Instability is measured by the average standard deviation of the errors across the 100 rotations for each (free) atom.  Smaller instability values indicate the model is closer to being rotation-covariant, where a fully rotation-covariant model would always a value of 0 regardless of its force MAE.
    }
    \label{tab:rot_analysis}
    \renewcommand{\arraystretch}{1.0}
    \resizebox{1.0\linewidth}{!}{
    \begin{tabular}{ccccccc}
      \toprule
         \mr{2}{\textbf{Dataset}} &  \textbf{Rotation}  & \textbf{Average instability of} & \mc{2}{c}{\textbf{Val Force MAE}} \\
        & \textbf{aug.} &  \textbf{per-atom force pred.} & \textbf{ID} & \textbf{Average} & \\
      \midrule
        All (130M) & \CheckmarkBold & \textbf{0.0037} & \textbf{0.0313} &  \textbf{0.0360} \\
        All (130M) & &  0.0069 & 0.0314 & 0.0366 \\
        2M & \CheckmarkBold & 0.0041  & 0.0332 & 0.0382 \\
        2M & & 0.0093 & 0.0346 & 0.0400 \\
      \bottomrule
    \end{tabular}}
    \vspace{-0.35cm}
\end{table}

\vspace{-0.2cm}
\subsection{S2F Performance Comparison}
We consider the S2F task (atomic force prediction) and compare the performance of \model{} against the baseline models, while normalizing for the computational time. Main results are plotted in Figure~\ref{fig:main_comp_efficiency}, and complete results can be found in Table~\ref{tab:comp}.

From Figure \ref{fig:main_comp_efficiency} (left), we see that \model{} gives superior force prediction performance given limited training GPU budgets.
Compared with SchNet, \model{} converges to a better performance with about 6.3 times less compute time. Compared to DimeNet++(resp.~-large), \model{}(resp.~-large) converges to the comparable performance with 18.9 (resp.~8.2) times less training compute. Compared to the GNS model, \model{} achieves much better performance across all training times although \model{} takes slightly longer to converge.

In Figure \ref{fig:main_comp_efficiency} (right), we see that 
\model{} achieves prediction performance comparable to DimeNet++, while enabling much faster inference speed. Specifically, compared to DimeNet++~(resp.~DimeNet++-large), \model{}++~(resp.~\model{}-large) is 6.5~(resp.~7.7) times faster, with comparable prediction performance. 

Finally, when \model{}-large is compared against DimeNet++, \model{}-large reduces the force prediction error by almost 10\% on average, while being 3.0 times faster in training and 2.4 times faster in inference.

\vspace{-0.2cm}
\subsection{Analysis of \model{}'s Rotation-covariance}
\label{subsec:rotcov}
We analyze whether \model{} is able to learn rotation-covariance when predicting atomic forces. We do so by measuring the prediction instability of \model{} when validation systems are rotated.

Table \ref{tab:rot_analysis} shows the results where all the models are trained to convergence for comparable times (training time for the 2M data is 80\% of training on all data).
We see a clear trend that both large data and rotation augmentation help to reduce the instability of \model{}'s prediction against rotation.
Moreover, the instability of \model{} trained on all data and rotation augmentation is relatively small compared to its force MAE, suggesting that \model{}'s prediction is close to be rotation covariant in the practical sense.

\vspace{-0.2cm}
\subsection{Ablation on \model{}'s model designs}
Here we perform extensive ablation studies on \model{}'s key design choices presented in Section~\ref{sec:model}.

\subsubsection{Basis and activation functions}
\label{subsubsec:basis}
We systematically study how choices of different basis and non-linear activation functions affect the model performance. We also compare with the ``None'' baseline, which does not use a basis function and directly concatenates the input raw edge feature into the node embeddings. 
In Figure \ref{fig:basis}, we see that the combination of spherical basis and Swish activation performs the best. 
For comparison, the GNS model uses no basis function (``None'') and ReLU, see Appendix \ref{app:gns} for full GNS model details. 

\subsubsection{Architecture design}
Next, we study the architectural building blocks of our conditional-filter-based message passing with the fixed basis and activation functions. We consider seven cases: 
{\bf (1) Only-dist}: we remove $\bm{n}_{st}$ from the input edge feature, \ie, $\bm{e}_{st} \equiv \bm{p}_{st}$, resulting in the edge features being rotation invariant.  
{\bf (2) No-atomic-radii}: we set the input edge features to $\bm{e}_{st} \equiv {\rm Concat} \left(\bm{n}_{st}, \|\bm{d}_{st} \| \right)$ (atomic radii information is dropped), 
{\bf (3)  No-node-emb}: filter ${\bm F}_c$ is a function of only $\bm{e}_{st}$ (conditioning on source and target node embeddings $\bm{h}_s^{(k)},\bm{h}_t^{(k)}$ is dropped), 
{\bf (4) Only-${\bm F}_c$}: Filter is directly aggregated, \ie, $\bm{m}_{st} = \bm{F}_c$, and self-message $m_t$ is omitted.
{\bf (5)  Edge-linear-BN}: MLP $\bm{F}_{\rm e}$ is replaced with a linear function followed by batch normalization,
{\bf (6) Node-linear-BN}: MLP $\bm{F}_{\rm n}$ is replaced with a linear function followed by batch normalization. 
{\bf (7) No-$\bm{m}_t$}: self-message $\bm{m}_t$ is removed.
Note that in {\bf (5)} and especially {\bf (6)}, we find it critical to add the batch normalization to facilitate training.

Table \ref{tab:ablation_message} shows the results of the seven ablation studies. Most notably, {\bf (1)} is significantly worse than the rest, because rotation-invariant node embeddings are insufficient for predicting rotation-covariant forces. We also see from {\bf (2)} and {\bf (3)} that making the filter less expressive, especially by dropping the dependency on node embeddings, significantly hurts performance. The improvement from element-wise product parameterization ${\bm F}_c \odot {\bm F}_d$ is demonstrated in {\bf (4)}. From {\bf (5)}, we see that it is critical to utilize non-linear models for edge features, as atomic forces are highly dependent on their subtle changes, but non-linearities are not essential for node embeddings {\bf (6)}. 
Finally, from {\bf (7)}, we see that the self-message $\bm{m}_t$ is not essential in performance.

Overall, our analysis suggests that \model{} benefits most from its expressive \emph{edge-level} computation via the conditional filter, which is directly responsible for accurately encoding the $3$D neighborhood structure on which the atomic forces depend.

\subsubsection{Model Scaling}
Comparing \model{} and \model{}-large in Table~\ref{tab:comp}, we see that a larger model provides significant performance gain, at the cost of 6.3 times more training time and 2.4 times more inference time. Extrapolating through Figure~\ref{fig:main_comp_efficiency}, we expect \model{} to significantly outperform DimeNet++, once comparable computational resources are used. We leave this investigation to future work.
More fine-grained ablations on model scaling are shown in Table~\ref{tab:scale}. We see that all the three scaling components help in the current regime of \model{}.


\section{Conclusions}
\label{sec:conclusion}



In this work, we demonstrate that force-centric GNN models without any explicit physical constraints are able to predict atomic forces more accurately than state-of-the-art energy-centric GNN models, while being faster both in training and inference.
We achieve this by carefully designing the message passing architecture, and by training the models on massive data and applying physics-based data augmentation.

This work opens up numerous avenues for future research: (1) Apply the same principle to other prediction tasks in OC20 (\eg, predicting per-system energy) and other application domains.
(2) Incorporate physics knowledge into \model{} to increase its generalization performance. 
(3) Improve computational efficiency of \model{}, while maintaining its performance, similar to how DimeNet++ has been significantly improved over DimeNet~\citep{klicpera_dimenetpp_2020}.
(4) Scale up \model{} with more computational resources.

\section*{Acknowledgements}
This work was done while Weihua Hu and Muhammed Shuaibi were at Facebook AI Research.
We acknowledge Pytorch~\citep{paszke2019pytorch} and Pytorch Geometric~\citep{fey2019fast}.
We thank Ryotatsu Yanagimoto and Zack Ulissi for insightful discussions on physics and chemistry.

We also gratefully acknowledge the support of
DARPA under Nos. N660011924033 (MCS);
ARO under Nos. W911NF-16-1-0342 (MURI), W911NF-16-1-0171 (DURIP);
NSF under Nos. OAC-1835598 (CINES), OAC-1934578 (HDR), CCF-1918940 (Expeditions), IIS-2030477 (RAPID);
Stanford Data Science Initiative, 
Wu Tsai Neurosciences Institute,
Chan Zuckerberg Biohub,
Amazon, JPMorgan Chase, Docomo, Hitachi, JD.com, KDDI, NVIDIA, Dell, Toshiba, and UnitedHealth Group. 
Weihua Hu is supported by Funai Overseas Scholarship and Masason Foundation Fellowship.
Jure Leskovec is a Chan Zuckerberg Biohub investigator.

\bibliography{reference}
\bibliographystyle{icml2021}

\clearpage
\appendix
\begin{center}
  \textbf{\Large Supplementary Material}
\end{center}




\section{Description of \dataset{} dataset}
\label{app:dataset}
\begin{figure*}
	\begin{center}
		\includegraphics[width=0.88\linewidth]{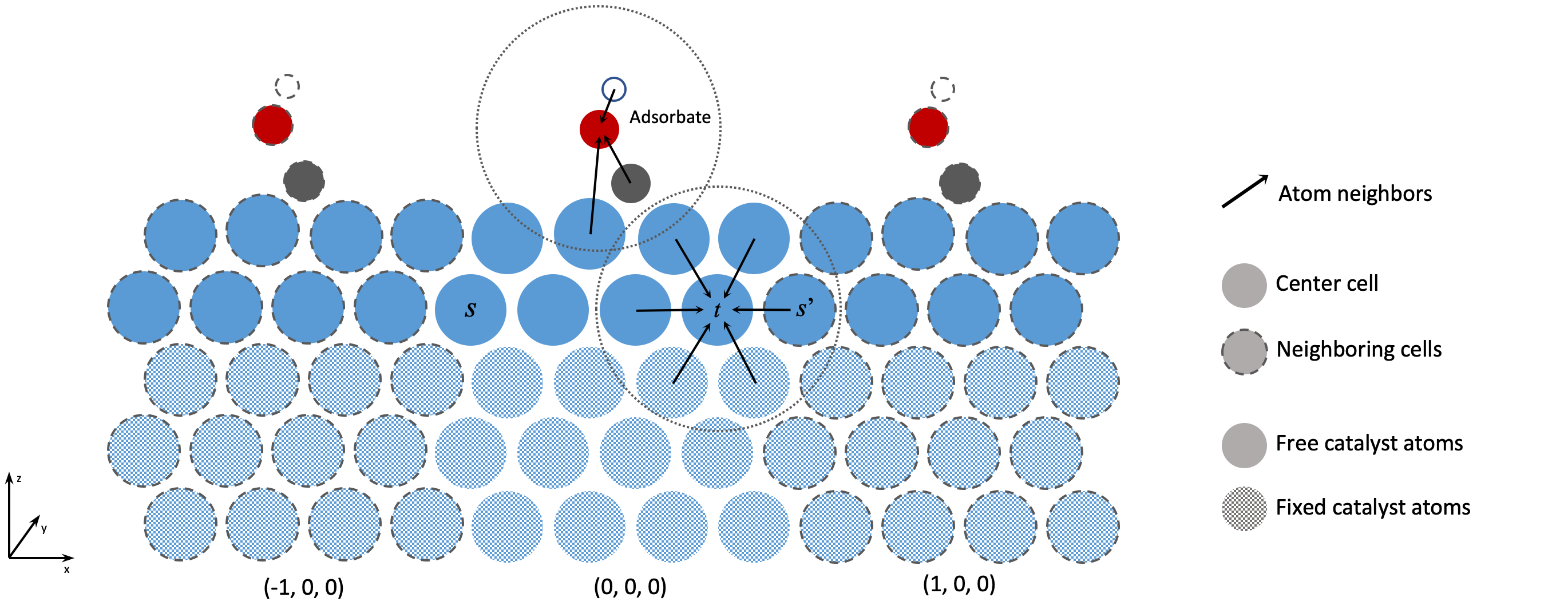}
	\end{center}
	\caption{2D Illustration of a slab that represents a catalyst's surface and an adsorbate. The slab is tiled in the $x$ and $y$ directions to create the surface (neighboring cells shown as atoms with dashed outlines). Only the cells to the left ($[-1, 0, 0]$) and right ($[1, 0, 0]$) are shown. The adsorbate is also assumed to be tiled with the slab (white, red, and grey atoms). Only the top 2 layers of the slab are allowed to move during a relaxation (dark blue), and the others are fixed (light blue). Neighboring atoms (black arrows) can be from the same cell or neighboring cells ($t$ and $s'$). All atoms within a radius (dotted circle) are assumed to be neighbors.}
	\label{fig:slab}
\end{figure*}

The \ocd{} dataset~\citep{OC20} contains over 130M non-equilibrium structures for training for the S2F task (\ie, atomic force prediction). The structures come from over 650K relaxation trajectories---the movement of the atoms from the initial structure to relaxed structure (equilibrium 3D structures with all-zero atomic forces).

Each structure contains the 3D positions of atoms in an adsorbate and catalyst slab, Figure \ref{fig:slab}. The adsorbate is a molecule involved in the chemical reaction that interacts with the catalyst's surface.  The adsorbate contains 1 to 11 atoms. The catalyst is represented as a ``slab'' that repeats infinitely in the $x$ and $y$ directions. The slab structure is repeated in a grid pattern where each repetition is referred to as a ``cell''. The center cell has coordinate $(0,0,0)$ with the cell to the left right being $(-1,0,0)$ and $(1,0,0)$ respectively. The slab is not repeated in the $z$ direction. Instead, the atoms at the bottom of the slab are assumed to be fixed and not move during a relaxation, which approximates how they would be held in place by the catalyst's atoms below the slab. Typically, only the top two layers of the catalyst's surface are assumed to be free and are moved according to their forces during a relaxation (see Figure \ref{fig:slab}). Therefore, forces are only evaluated on free catalyst atoms and the adsorbate.

The forces on the same atom in different cells are identical, since their atom neighbors are identical, resulting in their GNN's node embeddings to be also identical, \eg, the node embedding of atoms marked $s$ and $s'$ in Figure \ref{fig:slab} are the same. 
When computing the neighborhood of an atom, atoms in neighboring cells need to be also taken into consideration (atom $t$ in Figure \ref{fig:slab}). 
Notice that the message from $s$ to $t$ and the message from $s'$ to $t$ are different since the the relative placements of the two atoms are different, resulting in different edge features $\bm{e}_{st}$ and $\bm{e}_{s't}$. Computing the edge features between atoms from different cells can be done using the supplied information in the \ocd{} dataset for periodic boundary conditions.

\section{Details of GNS Model}
\label{app:gns}
For the GNS results in this paper, we reimplemented the original GNS model~\citep{sanchez2020learning}. Since the public code for the original GNS model~\citep{sanchez2020learning} was not available at the time of our experiments, we communicated with one of the authors to confirm the implementation details.

The message in the GNS model is defined as 
\begin{align*}
    m(\bm{h}_t^{(l)}, \bm{e}_{st}, \bm{h}_s^{(l)}) = \bm{{\rm MLP}}\left( {\rm Concat}\left(\bm{h}_t^{(l)}, \bm{e}_{st}, \bm{h}_s^{(l)} \right) \right) ,    
\end{align*}
where $\bm{{\rm MLP}}(\cdot)$ is a 1-hidden-layer MLP with ReLU activation and layer normalization~\citep{ba2016layer} applied before the activation.
For aggregating the message, the GNS model used either mean or sum, so we tried both in our experiments. We found sum aggregation to perform better, and report results of sum aggregation in this paper.
After the messages are aggregated, GNS uses a learnable linear function to transform the node embeddings.
Similar to \model{}, we additionally apply a batch normalization on the node embeddings, which alleviates training instability and significantly improves performance.
The GNS model uses a residual connection, where the computed node embeddings are added into the node embeddings from the previous layer. For the decoder, the GNS model uses a 1-hidden-layer MLP with ReLU activation. All the node embeddings and hidden units in the MLPs have the same dimensionality.

\section{Hyper-parameters}
\label{app:hyperparam}
For training, we use the Adam optimizer~\citep{kingma2014adam}, with an initial learning rate of 0.0005. 
We train \model{} and the GNS model for 500K iterations with the batch size of 256, which is equivalent to 1 epoch for the entire dataset\footnote{We do not observe much gain by training models longer than 1 epoch. This is probably because of the redundancy in data, \ie, out of 130M data points, there are 650k unique atom configurations (ignoring the positional differences).}. 
For \model{}-large, we use the batch size of 512.
All the parameters of the force-centric models are initialized with Xavier uniform initialization~\citep{glorot2010understanding}. The learning rate is kept constant for the first 250K iterations, after which it is halved every 50K iterations. 
We use the checkpoint with the best validation ID performance, and evaluate the saved model over all four validation sets. MAE over forces is used as the training loss. We will evaluate our models on the hidden test sets once the test server is ready.

For Gaussian and sine basis functions, we use $J=50$, which gives an output dimensionality of $B=350$. For Linear+Act, we set $B=350$. For spherical basis, we use $L=3$ and $S=4$, which results in $B = 36 (=3^2 \cdot 4)$, and we set $J=50$ for the internally-used sine basis function. For encoding the input atomic node features, we first normalize each dimension to lie between 0 and 1, and adopt the same basis function as used for encoding the edge features. The exception is spherical basis that is specialized for 3D spaces, in which case, the sine basis is used to encode the input atomic node features. We find that increasing $J$ and $L$ beyond the above values does not improve the performance, while significantly decreasing them worsens the performance. 

\section{Full Ablation Results}
\label{app:ablation}
\begin{table}[t]
    \centering
    \caption{Ablations on basis and activation functions in the \model{} architecture.}
    \label{tab:actbasis}
    \renewcommand{\arraystretch}{1.0}
     \resizebox{1.0\linewidth}{!}{
    \begin{tabular}{llcccccc}
      \toprule
         \mr{2}{\textbf{Basis}} & \mr{2}{\textbf{Act.}} &  \mc{5}{c}{\textbf{Validation Force MAE (eV/$\ang$)}} \\
        & &  ID & OOD Ads. & OOD Cat. & OOD Both & {\bf Average} \\
      \midrule
        Spherical  & ReLU & 0.0324  & 0.0367 & 0.0346 & 0.0454 & 0.0373  \\
        {\bf Spherical}  & {\bf Swish}  & {\bf 0.0313} & {\bf 0.0355} & {\bf 0.0334} & {\bf 0.0439} & {\bf 0.0360} \\
        \midrule
        Sine & ReLU & 0.0324 &  0.0367 & 0.0346 & 0.0456 & 0.0374 \\
        Sine & Swish & {\bf 0.0317} & {\bf 0.0360} & 0.0342 & 0.0448 & 0.0367 \\
        \midrule
        Gauss & ReLU & 0.0335 & 0.0384 & 0.0359 & 0.0476 & 0.0389 \\
        Gauss & Swish & {\bf 0.0318} & 0.0364 & 0.0346 & 0.0456 & 0.0371\\
        \midrule
        Linear+Act  & ReLU &  0.0340 & 0.0379 & 0.0356 & 0.0464 & 0.0385\\
        Linear+Act  & Swish & 0.0321  & {\bf 0.0359} & 0.0342 & 0.0445 & 0.0367 \\
        \midrule
        Identity  & ReLU & 0.0335 & 0.0377  & 0.0353 & 0.0464 & 0.0382 \\
        Identity  & Swish & 0.0322 & 0.0364 & {\bf 0.0338} & 0.0445 & 0.0368\\
        \midrule
        None  & ReLU &  0.0379  & 0.0430 & 0.0391 & 0.0519 & 0.0430 \\
        None  & Swish &  0.0330 & 0.0383 & 0.0347 & 0.0466 & 0.0382 \\
      \bottomrule
    \end{tabular}}
\end{table}

\begin{table}[t]
    \centering
    \caption{Ablations on the message passing architecture of \model{}.}
    \label{tab:archi}
    \renewcommand{\arraystretch}{1.0}
    \resizebox{1.0\linewidth}{!}{
    \begin{tabular}{lccccc}
      \toprule
          \mr{2}{\textbf{Ablation}} &  \mc{5}{c}{\textbf{Validation Force MAE (eV/$\ang$)}} \\
         &  ID & OOD Ads. & OOD Cat. & OOD Both & {\bf Average}  \\
      \midrule
         {\bf \model{}}  & {\bf 0.0313} & {\bf 0.0355} & {\bf 0.0334} & {\bf 0.0439} & {\bf 0.0360} \\
         (1) Only-dist & 0.0658 & 0.0673 & 0.0660 & 0.0805 & 0.0699 \\
         (2) No-atomic-radii & 0.0321 & 0.0362 & 0.0342 & 0.0447 & 0.0368 \\
         (3) No-node-emb & 0.0361 & 0.0409 & 0.0374 & 0.0495 & 0.0410 \\
         (4) Only-${\bm F}_c$ & 0.0333 & 0.0372 & 0.0350 & 0.0455 & 0.0378 \\
         (5) Edge-linear-BN & 0.0371 & 0.0430 &  0.0388 & 0.0520 & 0.0427 \\
         (6) Node-linear-BN & {\bf 0.0317} & {\bf 0.0356} &  {\bf 0.0339} & {\bf 0.0442} & {\bf 0.0364} \\
         (7) No-${\bm m}_t$ & {\bf 0.0314} & {\bf 0.0360} & {\bf 0.0336} & {\bf 0.0444} & {\bf 0.0364} \\
      \bottomrule
    \end{tabular}}
\end{table}

Here we provide full S2F results of our ablation studies, reporting the force MAE on each of the four validation sets.

\paragraph{Full Results on \model{} Designs}
First, we provide the full ablation results on \model{} designs.
Table \ref{tab:actbasis} shows the ablations on basis and activation functions, while Table \ref{tab:archi} shows the ablations on the message passing architectures.

Overall, we see trends that are consistent with the averaged results in Figure \ref{fig:basis} and Table \ref{tab:ablation_message}.
Specifically, from Table \ref{tab:actbasis}, we see that the combination of spherical basis functions and the Swish activation results in the best performance across the four validation sets.
From Table \ref{tab:archi}, we see that the conditional filter convolution design gives superior performance compared to the more simplified architectures, except for (6) and (7), in which the performance is comparable.

\paragraph{Full Results on Training Strategies}
Next, we provide the full ablation results of our training strategies, fixing the model architecture to the default \model{}.

The results are shown in Table \ref{tab:train_strategy}. 
First, we see that rotation augmentation helps, especially for the three out-of-distribution validation sets. Second, we see that providing small reweighted supervision on fixed atoms is also helpful, significantly improving the validation performance (evaluated on \emph{free atoms}) compared to the two baseline strategies:
(1) uniform loss weighting (equally weighting the losses for fixed and free atoms) and (2) zero-loss weighting (ignoring losses on fixed atoms during training).

\begin{table}[t]
    \centering
    \caption{Ablations on training strategies for \model{}.}
    \label{tab:train_strategy}
    \renewcommand{\arraystretch}{1.0}
    \resizebox{1.0\linewidth}{!}{
    \begin{tabular}{lclcccccc}
      \toprule
         \mr{2}{\textbf{Model}} &  \textbf{Rotation} & \textbf{Weight on} &  \mc{5}{c}{\textbf{Validation Force MAE (eV/$\ang$)}} \\
        & \textbf{aug.} & \textbf{fixed atoms} & ID & OOD Ads. & OOS Cat. & OOD Both & {\bf Average}  \\
      \midrule
        \model  & \CheckmarkBold & 0.05 & {\bf 0.0313} & {\bf 0.0355} & {\bf 0.0334} & {\bf 0.0439} &  {\bf 0.0360} \\
        \model  & & 0.05 & {\bf 0.0314} &{\bf 0.0359} & 0.0341 & 0.0448 & 0.0366 \\
        \model  & \CheckmarkBold & 1 & 0.0369 & 0.0411 & 0.0390 & 0.0506 & 0.0419 \\
        \model  & \CheckmarkBold & 0 & 0.0333 & 0.0385 & 0.0348 & 0.0465 & 0.0383 \\
      \bottomrule
    \end{tabular}}
\end{table}

\section{Structure Relaxation Simulation Results}
\label{app:IS2RS}
Here we apply \model{} to the IS2RS (\textbf{I}nitial \textbf{S}tructure to \textbf{R}elaxed \textbf{S}tructure) task. The goal is to predict the relaxed structure, \ie, 3D structure with zero-forces on all free atoms, from the initial structure. 
This can be achieved by simulating a relaxation trajectory: iteratively updating the atomic positions of free atoms according to their predicted forces until convergence, \ie, the predicted forces are below a pre-specified threshold.

Specifically, the structure relaxations are performed using a PyTorch implementation of the Atomic Simulation Environment's (\textit{ASE}) \citep{HjorthLarsen2017} L-BFGS optimizer. Relaxations were terminated when a max-absolute per-atom force of 0.01 eV/$\ang$ or 200 simulation steps, whichever comes first. All DFT calculations were performed in the \textit{Vienna Ab Initio Simulation Package} (VASP) \citep{Kresse1994, Kresse1996, Kresse1996a}. Both ASE and VASP are popular packages within the computational chemistry and catalysis communities.

The performance on the IS2RS task is evaluated by the two standard metrics~\citep{OC20}: (1) Average Force below Threshold (AFbT), measuring whether the predicted relaxed structure actually has small forces calculated by ground-truth DFT, and (2) Average Distance within Threshold (ADwT), measuring the geometrical closeness between the predicted relaxed structure and ground-truth relaxed structure. For both metrics, the higher, the better.

\begin{figure*}
\centering
\includegraphics[width=0.95\linewidth]{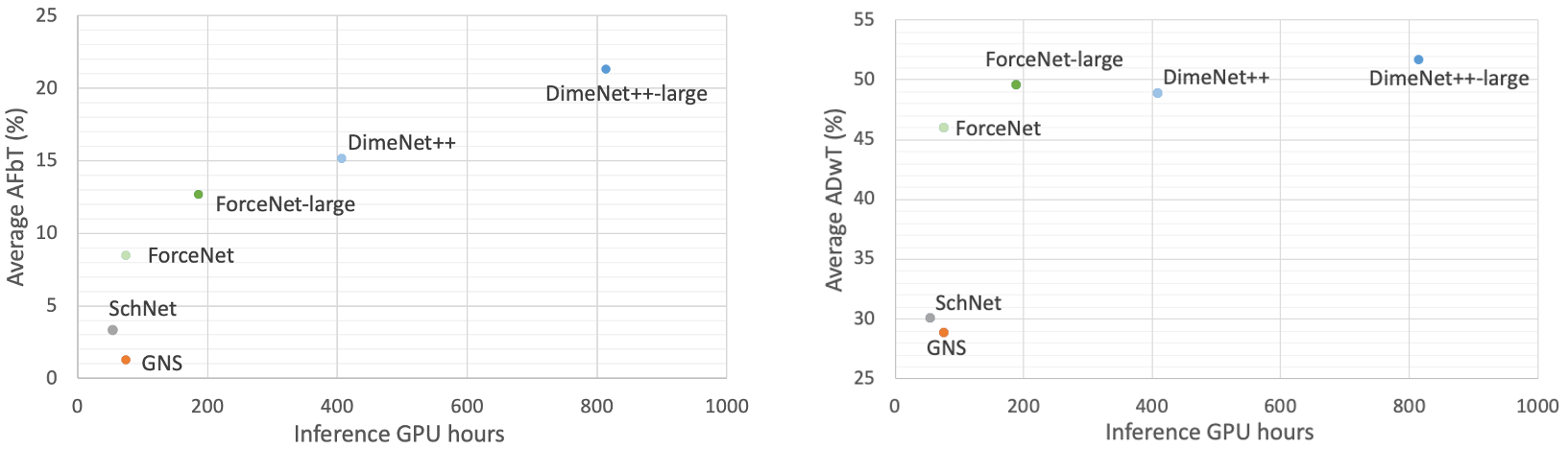}
\caption{Comparison of IS2RS performance in terms of AFbT and ADwT averaged over the four validation sets. The $x$-axis is the IS2RS inference time in GPU hours measured over 100K relaxations. }
\label{fig:is2rs}
\vspace{-0.3cm}
\end{figure*}

Figure~\ref{fig:is2rs} compares the performance of different models, while taking the inference efficiency into account.
The full results for all the validation sets are provided in Table~\ref{tab:is2rs_full}.
Here the inference time is measured on a Tesla V100 Volta GPU, where we use the largest possible batch size for each model and perform 100K relaxations. 
All the models are the same as Figure~\ref{fig:main_comp_efficiency} and Table~\ref{tab:comp}, originally trained for the S2F task.

We see from Figure~\ref{fig:is2rs} (left) that in terms of AFbT, both \model{} models outperform GNS and SchNet, while the inference time of \model{}, GNS and SchNet is comparable to each other. 
Compared to DimeNet++, both \model{} and \model{}-large have lower AFbT. However, the inference of both \model{} models is much faster than that of DimeNet++ (5.4 times faster for \model{}, and 2.2 times faster for \model{}-large).
Moreover, we see that there is still a room for \model{}-large to be further scaled up to give AFbT comparable to DimeNet++.
Regarding ADwT, from Figure~\ref{fig:is2rs} (right), we see that \model{}-large outperforms DimeNet++, while being 2.2 times faster in inference. \model{}-large is slightly worse than DimeNet++-large, but is 5.4 times faster in inference. 

Overall, the above results are encouraging given the faster inference time of \model{} compared to DimeNet++.
However, the results also suggest a potential limitation of \model{}'s force-centric approach: the superior performance of \model{}-large over DimeNet++ in the S2F task (\ie, estimate the forces of 3D structures along the simulation trajectory) does not directly translate into its superior performance on the IS2RS simulation task. 
We deduce this is due to the compounding error problem of the force-centric approach pointed out by the GNS work~\citep{sanchez2020learning}, \ie, model's prediction errors accumulate along the simulation trajectory, which forces the model to make increasingly erroneous prediction over structures that are far away from the simulation trajectory.
The energy-centric models may suffer less from the problem since their built-in physical constraints allow them to make more well-behaved force prediction over the off-trajectory structures, which eventually leads to better simulation results despite the worse force prediction results.

Fortunately, OC20 additionally provides 94M off-trajectory structures obtained by either perturbing the on-trajectory structures or performing molecular dynamics from relaxed structures~\citep{OC20}. 
We believe that these structures can be used to mitigate the compounding error problem of the force-centric approach by robustifying its off-trajectory force prediction.
In fact, the original GNS work~\citep{sanchez2020learning} has demonstrated that training their force-centric models on perturbed off-trajectory structures significantly reduces the compounding error, thereby improving their simulation results.
We leave this investigation to future work.

\begin{table*}[t]
    \centering
        \caption{Full IS2RS results. Inference time is in GPU hours and measured over 100K relaxations.}
    \label{tab:is2rs_full}
    \renewcommand{\arraystretch}{1.0}
    \setlength{\tabcolsep}{5pt}
    \resizebox{\linewidth}{!}{
    \begin{tabular}{lrrrrrr|rrrrr}
      \toprule
         \mr{2}{\textbf{Model}}  & \textbf{Inference} & \mc{5}{c}{\textbf{AFbT (\%)}} & \mc{5}{c}{\textbf{ADwT (\%)}} \\ 
        & \textbf{time} &  ID & OOD Ads. & OOD Cat. & OOD Both & {\bf Average} &  ID & OOD Ads. & OOD Cat. & OOD Both & {\bf Average}  \\
      \midrule
        GNS  & 74.3h & 2.22 & 0.66 & 1.44 & 0.62 & 1.24 & 30.60 & 23.13 & 30.92 & 31.15 & 28.95  \\
        SchNet & 54.1h & 4.90 & 2.66 & 2.75 & 2.90 & 3.30 & 35.54 & 29.80 & 26.86 & 28.39 & 30.15
        \\
        DimeNet++ & 407.6h & 17.41 & 14.41 & 14.19 & 14.55 & 15.14 & 48.75 &  45.19 & 48.59 & 53.14 & 48.92 \\
        DimeNet++-large & 814.6h & \textbf{24.22} & \textbf{20.40} & \textbf{20.13} & \textbf{20.31} & \textbf{21.27} & \textbf{52.45} & \textbf{48.47} & \textbf{50.98} & \textbf{54.82} & \textbf{51.68} \\
        \midrule
        \bf \model{} & 75.1h & 10.75 & 7.74 & 7.54 & 7.78 & 8.45 & 46.83 & 41.26 & 46.45 & 49.60 & 46.04   \\
      \bf \model{}-large & 186.9h & 14.77 & 12.23 & 12.16 & 11.46 & 12.66 & 50.59 & 45.16 & 49.80 & 52.94 & 49.62   \\
      \bottomrule
    \end{tabular}
    }
    \vspace{-0.3cm}
\end{table*}

\end{document}